\documentclass[11pt]{article}
\usepackage{coling2018}
\usepackage{times}
\usepackage{url}
\usepackage{latexsym}

\usepackage{times}
\usepackage{latexsym}
\usepackage{graphicx}
\usepackage{xcolor}
\usepackage{xspace}
\usepackage{latexsym}
\usepackage{tabularx}
\usepackage{microtype}
\usepackage{graphicx}
\usepackage{paralist}
\usepackage{scalefnt}
\usepackage{subfig}
\usepackage{calc}
\usepackage[utf8]{inputenc}
\usepackage{amsmath}
\usepackage{amssymb}
\usepackage{mathtools}
\usepackage{multirow}
\usepackage{booktabs}
\usepackage{paralist}

\usepackage{url}
\newcommand{\R}{\mathbb{R}}

\newcommand{\rt}[1]{\rotatebox{90}{#1}}

\newcommand{\ie}{\textit{i.\,e.}\xspace}
\newcommand{\eg}{\textit{e.\,g.}\xspace}

\newcommand{\F}{$\text{F}_1$\xspace}

\newcommand{\scl}{\textsc{Scl}\xspace}
\newcommand{\sda}{\textsc{Sda}\xspace}
\newcommand{\msda}{\textsc{mSDA}\xspace}
\newcommand{\nscl}{\textsc{Nscl}\xspace}
\newcommand{\noad}{\textsc{NoAd}\xspace}
\newcommand{\blse}{\textsc{Blse}\xspace}

\newcommand{\Csource}{C_{\textrm{source}}}
\newcommand{\Ctarget}{C_{\textrm{target}}}

\title{Projecting Embeddings for Domain Adaptation: \\
Joint Modeling of Sentiment Analysis in Diverse Domains}

\let\dagger1
\let\clubsuit2

\author{%
  Jeremy Barnes\textsuperscript{$\dagger$,$\clubsuit$},
  Roman Klinger\textsuperscript{$\dagger$}, \and
  Sabine Schulte im Walde\textsuperscript{$\dagger$} \\
  \textsuperscript{$\dagger$}Institut f\"ur Maschinelle Sprachverarbeitung\\ 
  University of Stuttgart\\
  Pfaffenwaldring 5b, 70569 Stuttgart, Germany\\
  \texttt{\{barnesjy,klinger,schulte\}@ims.uni-stuttgart.de}
  \AND
  \textsuperscript{$\clubsuit$}\rm{}Grup de Lingüística Computacional\\
  Universitat Pompeu Fabra\\ 
  Roc Boronat 138, 08018 Barcelona, Spain\\
  \texttt{jeremy.barnes@upf.edu}
}

\slname{Basque}

\date{}

\begin{document}
\maketitle

\begin{abstract}
  Domain adaptation for sentiment analysis is challenging due to the
  fact that supervised classifiers are very sensitive to changes in
  domain. The two most prominent approaches to this problem are
  structural correspondence learning and autoencoders. However, they
  either require long training times or suffer greatly on highly
  divergent domains. Inspired by recent advances in cross-lingual
  sentiment analysis, we provide a novel perspective and cast the
  domain adaptation problem as an embedding projection task. Our model
  takes as input two mono-domain embedding spaces and learns to
  project them to a bi-domain space, which is jointly optimized to (1)
  project across domains and to (2) predict sentiment. We perform
  domain adaptation experiments on 20 source-target domain pairs for
  sentiment classification and report novel state-of-the-art results
  on 11 domain pairs, including the Amazon domain adaptation datasets
  and SemEval 2013 and 2016 datasets.  Our analysis shows that our
  model performs comparably to state-of-the-art approaches on domains
  that are similar, while performing significantly better on highly
  divergent domains. Our code is available at
  \url{https://github.com/jbarnesspain/domain_blse}
\end{abstract}

\sltitle{Domeinu-Egokitzapenerako Bektore Proiekzioa:\\
         Domeinu Urrunetarako Sentimenduen Analisiko Eredu Bateratua}
\makesltitle
\begin{slabstract}
  Sentimenduen analisirako domeinu-egokitzapena erronka handi bat da
  oraindik, domeinu arteko ezberdintasunek ondorio esanguratsuak izan
  baititzakete sailkatzaile gainbegiratuentzat. Arazo honi aurre
  egiteko bi hurbilpen arrakastatsuenak egiturazko kidetasunaren
  ikasketa (structural correspondence learning) eta autoencoder-ak
  dira. Hala ere, denbora asko behar dute sistema entrenatzeko edo,
  domeinu arteko distantzia handia denean, ez dituzte emaitza onak
  lortzen. Hizkuntza-arteko sentimenduen analisian egindako azken
  lanetan oinarrituta, ikuspuntu berri bat eskaintzen dugu,
  domeinuaren egokitzapen ataza bektore proiekzio ataza gisa
  planteatuta. Gure sistemaren sarrera domeinu banako bi bektore
  espazio dira, zeinak sistemak espazio berri batera proiektatzen
  ikasten duen. Sistema hau optimizatuta dago (1) domeinu batetik
  besterako proiekzioa egiteko eta (2) esaldi baten sentimendua
  aurresateko. 20 jatorri-xede domeinu pareetan esperimentuak burutu
  ditugu eta 11 kasutan artearen egoerako emaitzarik onenak lortzen
  ditugu Amazon-eko domeinu-egokitzapeneko eta SemEval 2013 eta 2016
  datu-multzoetan. Gure analisian ikus daitekeenez, gure hurbilpena
  artearen egoerako sistemen pareko moldatzen da antzeko domeinuetan,
  baina emaitza hobeak lortzen ditu oso domeinu ezberdinetan. Gure
  kodea eskuragarri dago helbide honetan:
  \url{https://github.com/jbarnesspain/domain_blse}.
\end{slabstract}
\blfootnote{
  \hspace{-0.65cm}
  This work is licenced under a Creative Commons 
  Attribution 4.0 International Licence.
  Licence details:
  \url{http://creativecommons.org/licenses/by/4.0/}
}

\section{Introduction}
One of the main limitations of current approaches to sentiment
analysis is that they are sensitive to differences in domain. This
leads to classifiers that, after training, perform poorly on new
domains \cite{PangLee2008,Deriu2017}. Domain adaptation techniques
provide a solution to reduce the discrepancy and enable models to
perform well across multiple domains \cite{Blitzer2007}. The two main
approaches to domain adaptation for sentiment analysis are
\textit{pivot-based methods} \cite{Blitzer2007,Pan2010,Yu2016}, which
augment the feature space with domain-independent features learned on
unsupervised data, and \textit{autoencoder} approaches
\cite{Glorot2011,Chen2012}, which seek to create a good general
mapping from a sentence to a latent hidden space. While pivot-based
domain adaptation methods are well-motivated, they are often
outperformed by autoencoder methods. However, both approaches to
domain adaptation effectively lead to a loss of information, as they
must reduce the effect of discriminant features which are
domain-dependent. We argue in this paper that this leads to a
decreased performance especially in cases where the similarity between
the domains is low.

Unlike previous approaches, in this paper, we propose a domain
adaptation approach based on lessons learned from cross-lingual
sentiment analysis \cite{Barnes2018a}.  This approach maintains the
domain-dependent features, while adapting them to the target domain.
Following state-of-the-art approaches to create bilingual word
embeddings \cite{Mikolov2013,Artetxe2016,Artetxe2017}, we learn to
project a mapping from a source domain vector space to the target
domain space, while jointly training a sentiment classifier for the
source domain.

We show that our proposed model (1) performs comparably to
state-of-the-art models when domains are similar and (2) outperforms
state-of-the-art models significantly on divergent domains. We report
novel state-of-the-art results on 11 domain pairs. We also contribute
a detailed error analysis and compare the effect of different
projection lexicons. Our code is available at
\url{https://github.com/jbarnesspain/domain_blse}.

\section{Related Work and Motivation}

Domain adaptation is an omnipresent challenge in natural language
processing. It has been applied for many tasks, such as part-of-speech
tagging \cite{Blitzer2006,Daume2007}, parsing
\cite{Blitzer2006,Finkel2009,Mcclosky2010}, or named entity
recognition \cite{Daume2007,Guo2009,Yu2015}. In the following, we
limit our review to adaptation techniques which have been applied to
sentiment analysis.

\subsection{Pivot-based Approaches}
\newcite{Blitzer2006} propose \emph{structural correspondence
  learning} (\scl), which introduces the concept of
\emph{pivots}. These are features that behave in the same way for
discriminative learning for both domains, \eg, \textit{good} or
\textit{terrible} for sentiment analysis. The intuition is that
non-pivot domain-dependent features, \eg, \textit{well-written} for
the book domain or \textit{reliable} for electronics, which are highly
correlated to a pivot should be treated the same by a sentiment
classifier.

\newcite{Blitzer2007} extend their \scl approach to sentiment analysis
and also create one of the benchmark datasets for domain adaptation in
sentiment analysis. They crawl between 4000 and 7000 product reviews
for each domain, and create balanced datasets of 1000 positive and
1000 negative reviews for four product types (books, DVD, electronics,
and kitchen appliances). The remaining reviews serve as unlabeled
training data for the \scl approach. For each pivot, they train a
binary classifier to predict the existence of the pivot from non-pivot
features. They then use these classifiers to create a
domain-independent representation of the data. The concatenation of
the original representation and the \scl representation are used to
train a classifier.

\newcite{Pan2010} also exploit the relationship between pivots and
non-pivots to span the domain gap, but use a graph-based approach to
cluster non-pivot features and augment the original feature
space. \newcite{Yu2016} learn sentence embeddings that are useful
across domains through multi-task learning. They jointly train a
convolutional recurrent neural network model to predict the sentiment
of source domain sentences while at the same time predicting the
presence of pivots. Finally, \newcite{Ziser2017} propose neural
structural correspondence learning (\nscl), which marries \scl and
autoencoder techniques by using a neural network to create a hidden
representation of a text, and then using this representation to
predict the existence of pivots.

\nscl is currently state of the art, but requires a careful choice of
pivot features and extensive hyper-parameter search to achieve the
best results.

\subsection{Autoencoder Approaches}
\newcite{Glorot2011} adopt a deep learning approach for domain
adaptation. They create lower-dimensional representations for their
data through the use of \emph{stacked denoising autoencoders} (\sda),
which are trained to reconstruct the original sentence from a
corrupted version. They then train a linear SVM on the original
feature space augmented with the hidden representations obtained from
the autoencoder.

\newcite{Chen2012} extend this work by proposing \emph{Marginalized
  Denoising Autoencoders} (\msda), which are more scalable thanks to a
series of linear transformations which are performed in closed-form,
with the non-linearity being applied afterwards. This leads to a
significant gain in speed, as well as the ability to include more
features from the original representations. Autoencoder models perform
better than earlier \scl models (excluding \nscl), but have the
disadvantages of being less interpretable, requiring long training
times, and only utilizing a small amount of the original feature
space.

\subsection{Domain Specific Word Representations}
A third approach is to create word representations that provide useful
features for multiple domains. \newcite{He2011} propose a joint
sentiment-topic model which uses pivots to change the topic-word
Dirichlet priors. \newcite{Bollegala2015} create domain-specific
embeddings for pivots and non-pivots with the constraint that the
pivot representations are similar across domains.

The work that is most similar to ours is that of
\newcite{Bollegala2014}. Their method learns to predict differences in
word distributions across domains by learning to project
lower-dimensional SVD representations of documents across
domains. Unlike our work, however, they learn the projection step
separately from the classification. They also only learn to project
the features that the two domains have in common, which implies
discarding information useful for classification. These approaches,
however, perform worse than \msda and \nscl.

\section{Projecting Representations}

\begin{figure*}
\centering
\includegraphics[width=5.5in]{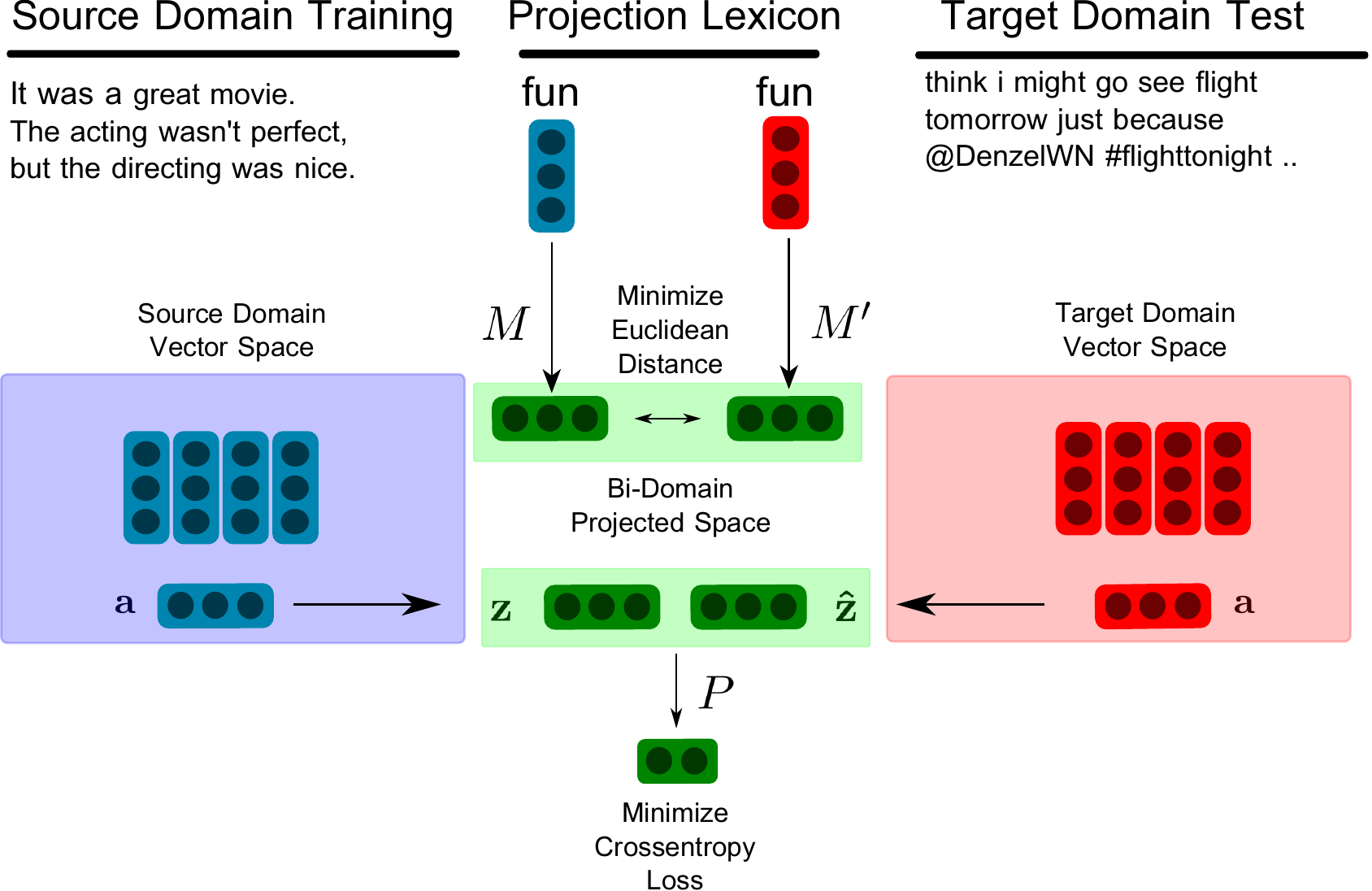}
\caption{Embedding projection architecture.}
\label{fig:model}
\end{figure*}

Our approach is motivated by previous success in learning to project
embeddings across languages for cross-lingual sentiment analysis,
namely \emph{Bilingual Sentiment Embeddings}
\cite[BLSE]{Barnes2018a}. The inputs for this model are (1)~a
monolingual embedding space for the source and target language, (2)~a
translation lexicon, and (3)~an annotated sentiment corpus for the
source language. It jointly learns to project the source and target
vectors into a bilingual space and also to classify the sentiment of
the sentences in the source corpus. At test time, the classifier is
able to use the projected target words as features because they are
optimized such that they resemble the source words.

In this work, we cast domain adaptation for sentiment analysis as a
version of this cross-lingual adaptation in which the source and
target domains have a large shared vocabulary. However, as is the case
in domain adaptation, words do not necessarily have the same semantics
across domains. Therefore, we will use the aforementioned projection
model to learn a word-level projection from one domain to another,
while jointly learning to classify the source domain. In the following, we
detail the projection objective, the sentiment objective, and the full
objective.

\subsection*{Cross-domain Projection}
\label{crossdomain}

We assume that we have two precomputed vector spaces $S = \R^{v \times
  d}$ and $T = \R^{v' \times d'}$ for our source and target domains,
where $v$ ($v'$) is the length of the source vocabulary (target
vocabulary) and $d$ ($d'$) is the dimensionality of the embeddings.
We also assume that we have a projection lexicon $L$ of length $n$
which consists of word-to-word pairs $L$ = $\{(s_{1},t_{1}),
(s_{2},t_{2}),\ldots, (s_{n}, t_{n})\}$ which map from source to
target domains. In this paper, we assume that the words map to
themselves across domains, so that $L$~=~$\{(s_{1}, s_{1}), (s_{2},
s_{2}),\ldots, (s_{n}, s_{n})\}$, although other mapping lexicons are
possible. One could imagine constructing a lexicon that maps concepts
from one domain to those of another, \ie ``read'' in the books domain
and ``watch'' for movies.

In order to create a mapping from both original vector spaces $S$ and
$T$ to shared sentiment-informed bi-domain spaces $\mathbf{z}$ and
$\mathbf{\hat{z}}$, we employ two linear projection matrices, $M$ and
$M'$. During training, for each translation pair in $L$, we first look
up their associated vectors, project them through their associated
projection matrix and finally minimize the mean squared error of the
two projected vectors. This is very similar to the approach taken by
\newcite{Mikolov2013}, but includes an additional target
projection matrix.

The projection quality is ensured by minimizing the mean squared
error\footnote{We omit parameters in equations for better readability.}
\begin{equation}
\textrm{MSE} = \dfrac{1}{n} \sum_{i=1}^{n} (\mathbf{z_{i}} - \mathbf{\hat{z}_{i}})^{2}\,,
\end{equation}
where $\mathbf{z_{i}} = S_{s_{i}} \cdot M$ is the dot product of the
embedding for source word $s_{i}$ and the source projection matrix and
$\mathbf{\hat{z}_{i}} = T_{t_{i}} \cdot M'$ is the same for the target
word $t_{i}$ and target matrix $M'$.

The intuition for including this second matrix is that a single
projection matrix does not support the transfer of sentiment
information from the source domain to the target domain. Although this
term is degenerate by itself, when coupled with the sentiment objective,
it allows the model to learn to project to a sentiment-aware target
language space.

\subsection*{Sentiment Classification}
\label{sentiment}

We add a second training objective to optimize the projected source
vectors to predict the sentiment of source phrases. This inevitably
changes the projection characteristics of the matrix $M$ and
consequently $M'$, which encourages $M'$ to learn to predict sentiment
without any training examples in the target domain.

To train $M$ to predict sentiment, we require a source-domain corpus
$\Csource = \{(x_{1}, y_{1}), (x_{2}, y_{2}), \ldots, (x_{i},
y_{i})\}$ where each sentence $x_{i}$ is associated with a label
$y_{i}$.

For classification, we use a multi-layer feed-forward architecture. 
For a sentence $x_{i}$, we take the word embeddings
from the source embedding $S$ and average them to $\mathbf{a}_{i} \in
\R^{d}$. We then project this vector to the joint bi-domain space
$\mathbf{z}_{i} = \mathbf{a}_{i} \cdot M$. Finally, we pass
$\mathbf{z}_{i}$ through a softmax layer $P$ to get our prediction
$\hat{y}_{i} = \textrm{softmax} ( \mathbf{z}_{i} \cdot P)$.

To train our model to predict sentiment, we minimize the cross-entropy
error of our predictions
\begin{equation} 
H = - \sum_{i=1}^{n} y_{i} \log \hat{y_{i}} - (1 - y_{i}) \log (1 - \hat{y_{i}})\,.
\end{equation}

\subsection*{Joint Learning}
\label{joint}

In order to jointly train both the projection component and the
sentiment component, we combine the two loss functions to optimize the
parameter matrices $M$, $M'$, and $P$ by
\begin{equation}
J =\kern-5mm
\sum_{(x,y) \in \Csource}\kern-1mm \sum_{(s,t) \in L}\kern-1mm \alpha H(x,y)
+ (1 - \alpha) \cdot \textrm{MSE}(s,t)\,,
\end{equation}
where $\alpha$ is a hyperparameter that weights sentiment loss vs.\ projection loss.

\subsection*{Target-domain Classification}
For inference, we classify sentences from a target-domain corpus
$\Ctarget$.
As in the training procedure, for each sentence, we take the word
embeddings from the target embeddings $T$ and average them to
$\mathbf{a}_{i} \in \R^{d}$. We then project this vector to the joint 
space $\mathbf{\hat{z}}_{i} = \mathbf{a}_{i} \cdot
M'$. Finally, we pass $\mathbf{\hat{z}}_{i}$ through a softmax layer
$P$ to get our prediction $\hat{y}_{i} = \textrm{softmax} (
\mathbf{\hat{z}}_{i} \cdot P)$.

\section{Experiments}

We compare our method with two adaptive baselines and one non-adaptive
version. We describe the six evaluation corpora in Section
\ref{datasets} and the baselines in Section \ref{baselines}.

\subsection{Datasets}
\label{datasets}

\newlength\mylength
\settowidth{\mylength}{*}
\newcommand{\mystar}{*\hspace{-\mylength}\mbox{}}

\begin{table}
\centering%\small
\begin{tabular}{lrrrrrrr}
\toprule
&    & \multicolumn{1}{c}{B} & \multicolumn{1}{c}{D}& \multicolumn{1}{c}{E} & \multicolumn{1}{c}{K} &\multicolumn{1}{c}{S13} & \multicolumn{1}{c}{S16}\\
\cmidrule(rl){2-2}\cmidrule(l){3-3}\cmidrule(l){4-4}\cmidrule(l){5-5}\cmidrule(l){6-6}\cmidrule(l){7-7}\cmidrule(l){8-8}
 \multirow{3}{*}{\rt{Train}}
& $+$   & 800 &800&800&800& 2,225 & 2,468 \\
& $-$   & 800 &800&800&800& 831 & 664\\
 \cmidrule(rl){2-2}\cmidrule(l){3-3}\cmidrule(l){4-4}\cmidrule(l){5-5}\cmidrule(l){6-6}\cmidrule(l){7-7}\cmidrule(l){8-8}
  \multirow{3}{*}{\rt{Dev}}
& $+$   & 200 & 200 &200   &200   & 328 & 682   \\
& $-$   & 200 & 200 &200   &200   & 163 & 310 \\
 \cmidrule(rl){2-2}\cmidrule(l){3-3}\cmidrule(l){4-4}\cmidrule(l){5-5}\cmidrule(l){6-6}\cmidrule(l){7-7}\cmidrule(l){8-8}
  \multirow{3}{*}{\rt{Test}}
& $+$   & 1000\mystar &1000\mystar  &1000\mystar   &1000\mystar   & 946 &  5,619 \\
& $-$   & 1000\mystar &1000\mystar  &1000\mystar   &1000\mystar   & 316 & 2,386\\
 \cmidrule(rl){2-2}\cmidrule(l){3-3}\cmidrule(l){4-4}\cmidrule(l){5-5}\cmidrule(l){6-6}\cmidrule(l){7-7}\cmidrule(l){8-8}
\textit{Total}  &  & 2,000 &2,000   &2,000     &2,000   & 4,809 & 12,129 \\
\bottomrule
\end{tabular}
\caption{Statistics for the Amazon corpora (books, DVD, electronics, kitchen), as well as the SemEval 2013 and 2016 message classification tasks (S13 and S16 respectively). * For the Amazon corpora, we test on the entire target domain corpora.}
\label{datasetstats}
\end{table}
\subsubsection{Amazon Corpora}

In order to evaluate our proposed method, we use the corpus collected
by \newcite{Blitzer2007}, which consists of Amazon product reviews
from four domains: books (B), DVD (D), electronics (E), and kitchen (K). Each
subcorpus contains a balanced labeled subset, with 1000 positive and
1000 negative reviews, as well as a much larger set of unlabeled
reviews. We use the standard split of 1600 reviews from each domain as
training data and the remaining 400 reviews as validation data. For
testing, we use all of the 2000 reviews from the target domain \cite{Ziser2017}.

We take the unlabeled data from each domain to create the domain
embeddings for our method, as well as to train the domain independent
representations for the \nscl and \msda methods. In order to create
embeddings for the Amazon corpora, we concatenate all of the unlabeled
data from all domains. The statistics for this corpus are given in
Table \ref{datasetstats}.

\subsubsection{SemEval Corpora}

Sentiment analysis of Twitter data is common nowadays, with several
popular shared tasks organized on the topic
\cite[\emph{i.\,a.}]{Nakov2013,TASS2013,Sentipol2014,Nakov2016}. In
order to evaluate how well domain adaptation techniques perform on
large domain gaps, we also use the message polarity classification
corpora provided by the organizers of SemEval 2013 and 2016
\cite{Nakov2013,Nakov2016}. We will refer to these as S13 and S16,
respectively. These contain tweets which have been annotated for
positive, negative, and neutral sentiment. We remove neutral tweets,
giving us a binary setup which allows compatibility with the Amazon
corpora. The statistics for these corpora are given in Table
\ref{datasetstats}.

\subsection{Embeddings}

For \blse, we create mono-domain embeddings using the Word2Vec
toolkit\footnote{\url{https://code.google.com/archive/p/word2vec/}} by 
training skip-gram embeddings with 300 dimensions, subsampling of
$10^{-4}$, window of 5, negative sampling of~15 on the concatenation
of the unlabeled Amazon corpora.
We also create Twitter-specific embeddings by training on nearly 8
million tokens taken from tweets collected using various hashtags. The
parameters were the same as those used to create the Amazon
embeddings.
For out-of-vocabulary words, a vector initialized randomly between
$-0.25$ and $0.25$ approximates the variance of the pretrained
vectors.

\subsection{Baselines and Model}
\label{baselines}
Domain transfer for sentiment analysis has been widely studied on the
Amazon sentiment domain corpus. However, we hypothesize that progress
previous approaches have made on this particular corpus may not hold
when tested on more divergent domains. Therefore, we compare two
state-of-the-art approaches on the Amazon corpus with our method, as
well as a standard non-adaptive baseline.

\textbf{\noad} is a non-adaptive approach which uses a bag-of-words
representation from each review as features for a linear SVM.

\textbf{\msda} is the original implementation of marginalized Stacked
Denoising Autoencoders \cite{Chen2012}, one of the state-of-the-art
domain adaptation methods on the Amazon sentiment domain corpus. The
approach learns a latent hidden representation of the data, which is
then concatenated to the original feature space. For our experiments,
we use the 30000 most common uni- and bi-grams as features and take
the top 5000 features as pivots \cite{Chen2012}. We tune the
corruption level (0.5, 0.6, 0.7, 0.8, 0.9) and the C-parameter for the
SVM classifier on the source domain validation data, but leave the
number of layers at 5.

\textbf{\nscl} is an approach that marries both the pivot-based
methods and autoencoders. Specifically, we use the original
implementation\footnote{\url{https://github.com/yftah89/Neural-SCL-Domain-Adaptation}}
of the Autoencoder SCL with similarity regularization, which we refer
to as \nscl. This approach substitutes the reconstruction weights of
the autoencoder with a matrix of the pre-trained word embeddings of
pivots. This enables the model to generalize beyond boolean
features. We set the hyper-parameters for training the autoencoders
with stochastic gradient descent to those from the original
paper\footnote{Learning rate: 0.1, momentum: 0.9, weight-decay
  regularization: $10^{-5}$} and tune the number of pivots (100, 200,
300, 400, 500), dimensionality of the hidden layer (100, 300, 500),
and C-parameter for logistic regression on the source domain
validation data (400 reviews).

\textbf{\blse} is our approach based on cross-domain vector
projection. We use the domain-specific word embeddings to initialize
our model and following the embedding literature, we take the most
common 20,000 words in the concatenated corpora as a projection
dictionary (see Section \ref{lexicon_choice}). We tune the
hyper-parameters training epochs, alpha (0.1--0.9), and batch sizes
(20--500) on the source domain validation data.

\subsection{Results}

Tables \ref{results:amazon} and \ref{results:semeval} present the
results of our experiments. In order to compare with previous work, we
report accuracy scores for the balanced Amazon corpora. Because the
SemEval corpora are highly imbalanced, we instead present macro \F
scores. We introduce the notation X$\rightarrow$Y, where X is the
train corpus and Y is the test corpus, to indicate the domain pairs.

On the Amazon corpora, Table~\ref{results:amazon}, \nscl outperforms
the other approaches (3.6 percentage points (pp) in \F on average
compared to \blse, 2.5\,pp compared to \msda, and 5.1\,pp compared to
\noad). \blse only performs better than \nscl on three setups (DVD to
books, electronics to DVD, and kitchen to DVD) and \msda on four
setups (DVD to books, books to DVD, electronics to DVD, and kitchen to
DVD). \blse performs better on the books and DVD test sets than the
electronics and kitchen test sets (an average of 3.23 pp). This can be
explained by the fact that the corpora used to train the Amazon
embeddings contain many more unlabeled reviews for books and
electronics (973,194 / 122,438 respectively) than electronics and
kitchen (21,009 / 17,856). Consequently, the vector representations
for sentiment words that only appear in the books and DVD subcorpora
are of higher quality than those that only appear in the electronics
and kitchen subcorpora (see Table \ref{DomainEmbeddings}). \blse
relies entirely on the embeddings as input, and if the quality of the
embeddings is lower, the model cannot use these features to correctly
classify a review. This suggests that the amount of available
unlabeled data in the target domain is important, but not limiting. In
this paper, we did not decide to crawl more data for the electronics
and kitchen domains, but this would be relatively straightforward.

\begin{figure*}
\begin{minipage}{0.45\textwidth}
\centering
\includegraphics[width = 3.2in]{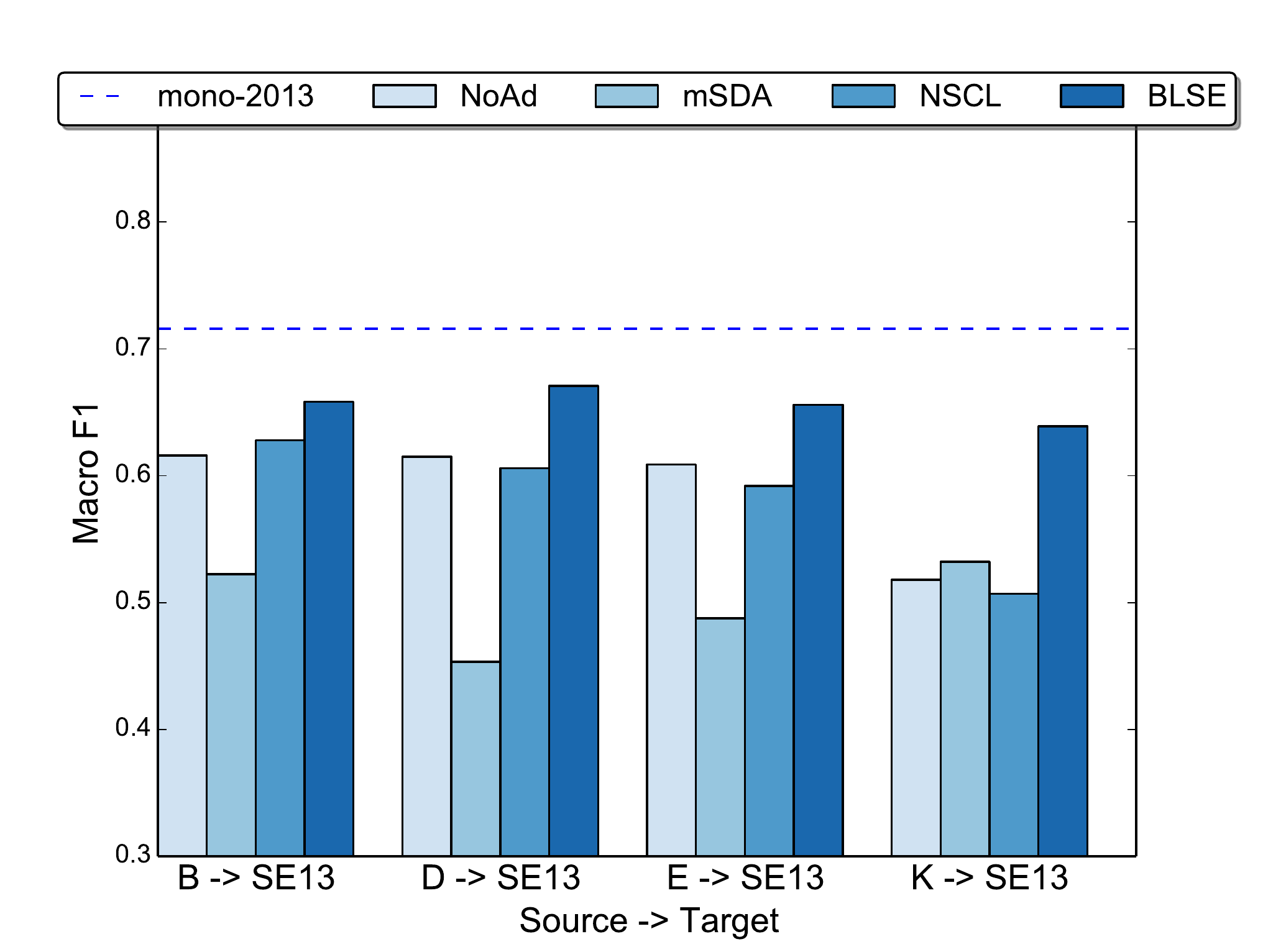}
\caption{\F of approaches trained on the source dataset and tested on the 2013 SemEval corpus.}
\label{fig:sem2013}
\end{minipage}
\hfill
\begin{minipage}{0.45\textwidth}
\centering
\includegraphics[width = 3.2in]{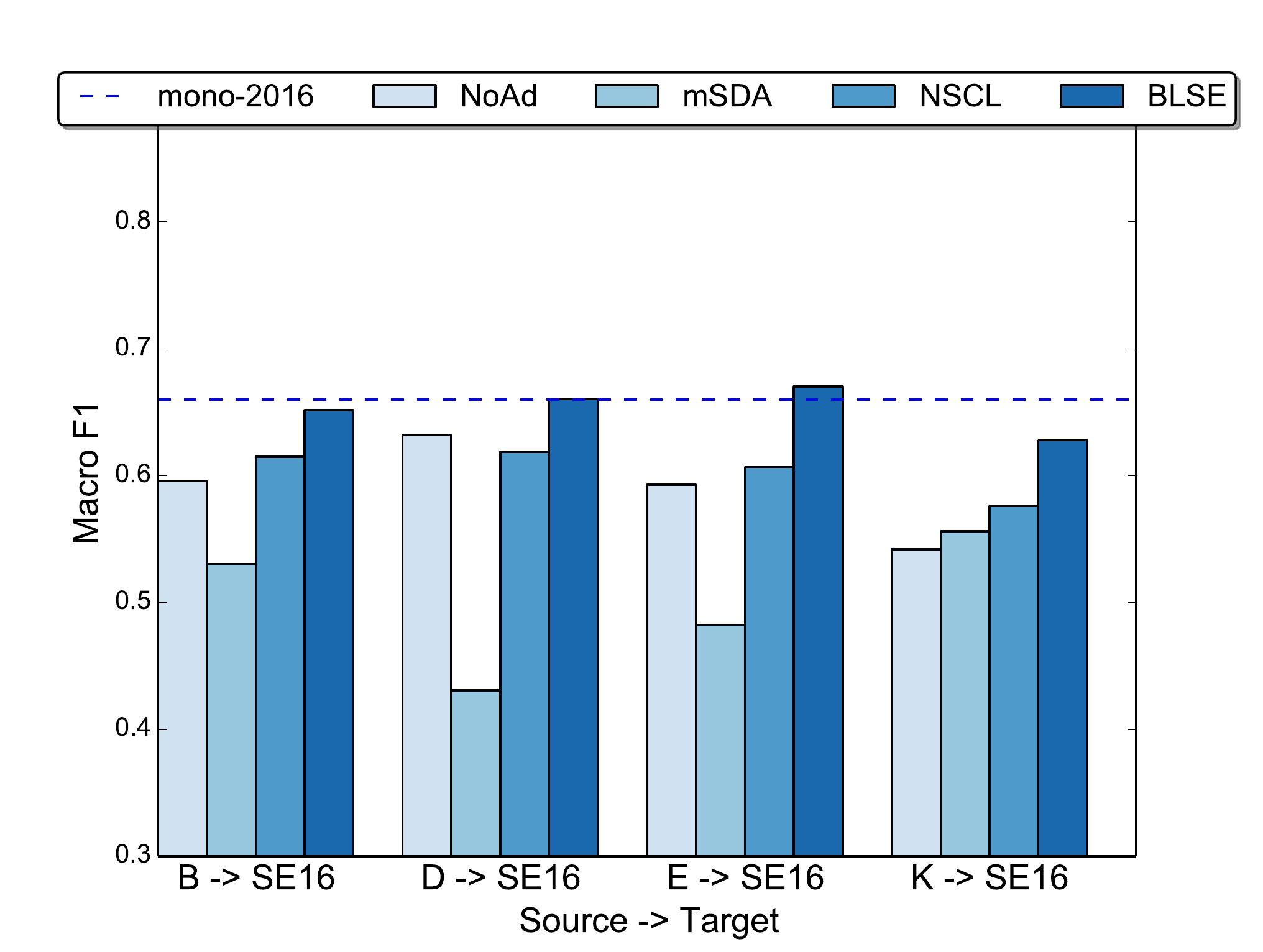}
\caption{\F of approaches trained on the source dataset and tested on the 2016 SemEval corpus.}
\label{fig:sem2016}
\end{minipage}
\end{figure*}

For the SemEval corpora (see Figures \ref{fig:sem2013} and
\ref{fig:sem2016}), \blse significantly outperforms all other models
(8.2, 15.5, and 6.4 \F better on average compared to \nscl, \msda, and
\noad, respectively). \nscl is better than \msda on 7 of the 8 setups,
but better than the \noad baseline on only 4. \msda performs
particularly poorly here and only outperforms the baseline on one
setup. We suspect that this may be caused by the substantial
differences in the source and target corpora and the way this affects
the representation given to the classifier, which we explore in more
detail in Sections \ref{divergence} and \ref{erroranalysis}.

\begin{table*}
\centering%\small
\setlength\tabcolsep{1.0mm}
 \begin{tabular}{c|ccc|ccc|ccc|ccc}
 \toprule
 & D$\rightarrow$B & E$\rightarrow$B & K$\rightarrow$B &B$\rightarrow$D & E$\rightarrow$D & K$\rightarrow$D & B$\rightarrow$E & D$\rightarrow$E & K$\rightarrow$E & B$\rightarrow$K & D$\rightarrow$K & E$\rightarrow$K\\ 
\cmidrule(rl){1-1}\cmidrule(lr){2-4}\cmidrule(lr){5-7}\cmidrule(lr){8-10}\cmidrule(lr){11-13}
\blse & \textbf{82.2} & 71.3 & 69.0 & 81.0 & \textbf{76.8} & \textbf{76.5} & 71.8 & 70.3 & 70.8 & 73.8 & 72.3 & 78.3 \\
\nscl & 77.3 & 71.2 & \textbf{73.0} & \textbf{81.1} & 74.5 & 76.3 & \textbf{76.8} & \textbf{78.1} & \textbf{84.0} & \textbf{80.1} & \textbf{80.3} & \textbf{84.6} \\
\msda & 76.1 & \textbf{71.9} & 70.0 & 78.3 & 71.0 & 71.4 & 74.6 & 75.0 & 82.4 & 78.8 & 77.4 & 84.5 \\
\noad & 73.6 & 67.9 & 67.7 & 76.0 & 69.2 & 70.2 & 70.0 & 70.9 & 81.6 & 74.0 & 73.2 & 82.4 \\
\bottomrule
 \end{tabular}
 \caption{Sentiment classification accuracy for the
   \newcite{Blitzer2007} task.}
 \label{results:amazon}
\end{table*}

\begin{table*}
\centering%\small
\setlength\tabcolsep{1.0mm}
 \begin{tabular}{c|cccc|cccc}
 \toprule
 & B$\rightarrow$S13 & D$\rightarrow$S13 & E$\rightarrow$S13 & K$\rightarrow$S13 &B$\rightarrow$S16 & D$\rightarrow$S16 & E$\rightarrow$S16 & K$\rightarrow$S16 \\ 
\cmidrule(rl){1-1}\cmidrule(lr){2-5}\cmidrule(lr){6-9}
\blse &\textbf{65.8} &\textbf{67.1} & \textbf{65.6} & \textbf{63.9} & \textbf{65.2} & \textbf{66.1} & \textbf{67.0} & \textbf{62.8}\\
\nscl & 62.8 & 60.6 & 59.2 & 50.7 & 61.5 & 61.9 & 60.7 & 57.6\\
\msda & 52.2 & 45.3 & 48.8 & 53.2 & 53.1 & 43.1 & 48.2 & 55.6\\
\noad & 61.6 & 61.5 & 60.9 & 51.8 & 59.6 & 63.2 & 59.3 & 54.2\\
\bottomrule
 \end{tabular}
 \caption{Sentiment classification macro \F for the SemEval 2013 and 2016 tasks in binary setup.}
 \label{results:semeval}
\end{table*}

\begin{table*}[t]
\centering%\small
 \begin{tabular}{crrrr}
 \toprule
& \multicolumn{2}{c}{books} &\multicolumn{2}{c}{electronics} \\
\cmidrule(lr){2-3}\cmidrule(l){4-5}
word & admires & conceit & indispensable & cumbersome \\
\cmidrule(r){1-1}\cmidrule(lr){2-3}\cmidrule(lr){4-5}
\multirow{4}{*}{\rt{neighbors}} & professes & conceits & career.this & choppiness \\
& unselfish & macgruffen & non-western & setups \\
& parminder & pretentiously & mindwalk & forgiveable \\
& well-liked & contrivance & all-too-rare & unweildy \\
\bottomrule
 \end{tabular}
 \caption{Words and their nearest neighbors for important domain-dependent sentiment words. The nearest neighbors for the two example words from the book domain are more coherent than those of the electronics domain.}
 \label{DomainEmbeddings}
\end{table*}

\begin{figure*}[t]
\centering
\includegraphics[scale=0.7]{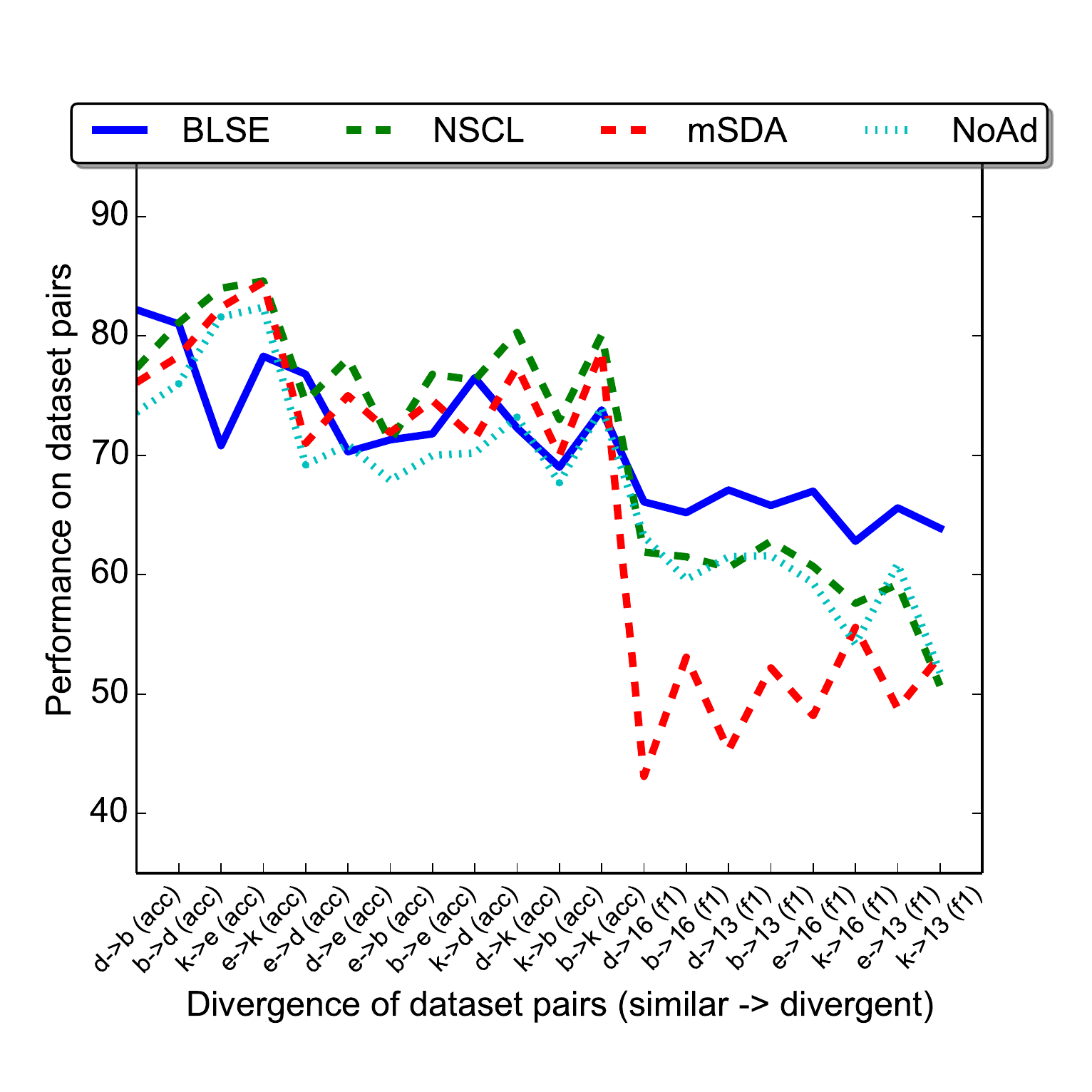}
\caption{Plot of performance of each model as a function of domain
  similarity. The x axis plots the rank of similarity from
  most similar (left) to least similar (right). \blse maintains its performance as the similarity
  decreases.}
\label{fig:overall}
\end{figure*}

\section{Model Analysis}
In this section we examine aspects of our model in an attempt to shed
light on its strengths and weaknesses. Specifically, we observe how
our model performs on highly divergent datasets, perform an error
analysis, motivate our choice of projection lexicon, and motivate the
need for $M'$.

\subsection{Domain Divergence and Feature Sparsity}
\label{divergence}

From the initial results, it seems that the \blse model performs
better on more divergent domains when compared to other
state-of-the-art models. In order to analyze this further, we test the
similarity of our domains using the Jensen-Shannon Divergence, which
is a smoothed, symmetric version of the Kullback-Leibler Divergence,
$D_{KL}(A||B) = \sum_{i}^{N} a_{i} \log
\frac{a_{i}}{b_{i}}$. Kullback-Leibler Divergence measures the
difference between the probability distributions $A$ and $B$, but is
undefined for any event $a_{i} \in A$ with zero probability, which is
common in term distributions. Jensen-Shannon Divergence is then
\[
D_{JS}(A,B) = \frac{1}{2} \Big[ D_{KL}(A||B) + D_{KL}(B||A)  \Big]\,.
\]
Our similarity features are probability distributions over terms $t
\in \mathbb{R}^{|V|}$, where $t_{i}$ is the probability of the $i$-th
word in the vocabulary $V$.

For each domain, we create frequency distributions of the most
frequent 10,000 unigrams that all domains have in common and measure
the divergence with $D_{JS}$. The results in Table
\ref{table:divergence} make it clear that the SemEval datasets are
more distant from the Amazon datasets than the Amazon datasets are
from each other. This is especially true for the distance between the
SemEval datasets from the kitchen dataset ($D_{JS}$ = 0.741 and 0.761,
respectively). This suggests that \nscl and \msda give the best
results when the difference between domains is relatively small,
whereas \blse performs better on more divergent datasets.

On the SemEval datasets, \blse also benefits from using dense
representations, rather than the sparse unigram and bigram features of
\nscl and \msda. This is particularly important when you have less
domain overlap and smaller texts (the average number of features for
the Amazon corpora is 76, compared to 17 for SemEval). \blse is always
able to find useful features, even if the tweet is quite short,
whereas a bag-of-words representation can be so sparse that it is not
helpful.

\begin{table*}[t]
\newcommand{\di}[1]{\underline{#1}}
\newcommand{\cl}[1]{\textbf{#1}}
\centering%\small
\begin{tabular}{lcccccc}
\toprule
             &   book &   DVD &   electronics &   kitchen &   SemEval 2013 &   SemEval 2016 \\
\cmidrule(rl){2-2}\cmidrule(l){3-3}\cmidrule(l){4-4}\cmidrule(l){5-5}\cmidrule(l){6-6}\cmidrule(l){7-7}
 book        &  1.000 & \cl{0.940} &         0.870 &     0.864 &         \di{0.775} &         0.802 \\
 DVD         &  	      & 1.000 &         0.873 &     0.866 &         \di{0.790} &         0.814 \\
 electronics &        &       &         1.000 &     \cl{0.908} &         \di{0.748} &         0.769 \\
 kitchen     &  && &     1.000 &         \di{0.741} &         0.761 \\
 SemEval 2013 & &&& &         1.000 &         \cl{0.921} \\
 SemEval 2016 & &&&& &         1.000 \\
\bottomrule
\end{tabular}
\caption{Jensen-Shannon divergence between term distribution
  representations of datasets. The \cl{bold} numbers represent the
  most similar domains and \di{underlined} numbers represent the most
  divergent.}
\label{table:divergence}
\end{table*}

\subsection{Error Analysis}
\label{erroranalysis}
We perform a label-based error analysis of the models on the SemEval
2013 and 2016 datasets by checking the error rate for the positive and
negative classes, which we define as
\begin{equation}
\textrm{Error Rate} = \dfrac{e_{c}}{n_{c} }\,,
\end{equation}
where the number of errors $e_{c}$ in class $c$ is divided by the
total number of examples $n_{c}$ in the class. We hypothesize that better
models will suffer less on minority classes. The results are found
in Table \ref{table:erroranalysis}. In general, \blse has better overall
performance than \nscl or \msda. In fact, \msda performs very poorly
on the minority negative class, with error rates reaching 98
percent. \nscl almost always favors a single class, with error rates
as high as 60.4 on negative and 70.5 on positive.

\subsection{Choice of Projection Lexicon}
\label{lexicon_choice}

Given that the choice of projection lexicon is one of the key
parameters in the \blse model, we experiment with three approaches to
creating a projection lexicon and observe their effect on the books to
SemEval 2013 setup.

The \textbf{Most Frequent Source Words} are a common source of
projection lexicon in the multilingual embedding literature
\cite{Faruqui2014,Lazaridou2015}. For our experiment, we take the
20,000 most frequent tokens from the Brown corpus
\cite{Francis1979}. The hypothesis behind using a general corpus is
that a large general lexicon will provide more supervision than a
smaller task-specific lexicon. This should contribute to learning
accurate projection matrices $M$ and $M'$.

\textbf{Sentiment Lexicons} often contain domain-independent words
that convey sentiment. In our model, using a sentiment lexicon as a
translation dictionary is equivalent to the use of pivots in other
frameworks, as these are usually domain independent words with are
good predictors of sentiment. The hypothesis here is that a small
task-specific lexicon will help to learn a good projection for the
most discriminative words. For our experiment, we take the subset of
the sentiment lexicon from \newcite{HuandLiu2004} which is found in
the Amazon and SemEval corpora. The final version has 1130 words.

\textbf{Mutual Information Selected Pivots} have been shown to be a
good predictor of sentiment across domains
\cite{Blitzer2006,Pan2010,Ziser2017}. We experiment with using words
with the highest mutual information scores as a projection lexicon,
although this leads to smaller lexicons. Our hypothesis is that
specific source-target domain lexicons may provide a better projection
between the two specific domains. For each source and target domain
pair, we take unigrams and bigrams with high mutual information scores
that appear at least 10 times in both domains. The number of pivots
differs with each domain pair. The lowest number is 100 (DVD to
SemEval 2013) and the highest 955 (books to DVD), with an average of
470 per domain pair.

\begin{table}[t]
\begin{minipage}[b]{0.53\textwidth}
\centering
\includegraphics[width = 1.0\linewidth]{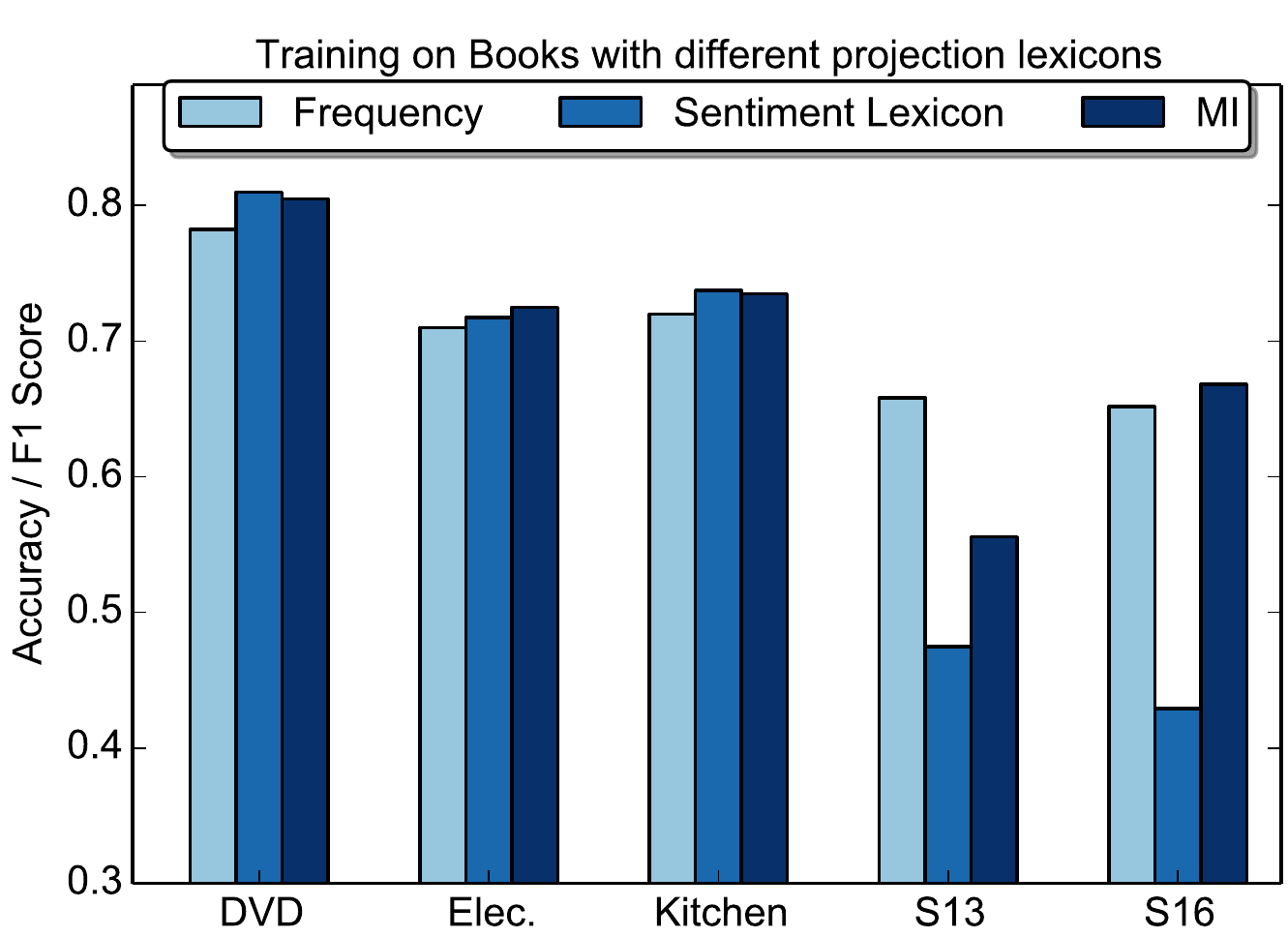}
\captionof{figure}{The effect of different projection lexicons for the \blse method when training on the books domain.}
\label{fig:lexicons}
\end{minipage}
\hfill
\begin{minipage}[b]{0.44\textwidth}
\centering\small
\begin{tabular}{llrr|rr}
\toprule
& & \multicolumn{2}{c}{SemEval 2013} & \multicolumn{2}{c}{SemEval 2016} \\
\cmidrule(lr){3-4}\cmidrule(lr){5-6}
& & Pos & Neg & Pos & Neg \\
\cmidrule(lr){3-4}\cmidrule(lr){5-6}
\multirow{3}{*}{\rt{books}}
& \blse & 26.9 & 35.4 & 31.8 & 33.0 \\
& \nscl & 34.2 & 43.0 & 28.9 & 43.5 \\
& \msda & 1.4  & 92.7 & 1.4 & 89.5 \\
\cmidrule(lr){3-4}\cmidrule(lr){5-6}
\multirow{3}{*}{\rt{DVD}}
& \blse & 22.8 & 39.2 & 28.0 & 36.5 \\
& \nscl & 18.1 & 60.4 & 18.7 & 52.1 \\
& \msda & 0.2 & 97.8 & 0.2 & 98.2 \\
\cmidrule(lr){3-4}\cmidrule(lr){5-6}
\multirow{3}{*}{\rt{elec.}}
& \blse & 19.1 & 48.7 & 27.7 & 34.6 \\
& \nscl & 35.2 & 41.5 & 38.9 & 33.7 \\
& \msda & 1.1 & 93.7 & 0.8 & 91.6 \\
\cmidrule(lr){3-4}\cmidrule(lr){5-6}
\multirow{3}{*}{\rt{kitchen}}
& \blse & 21.6 & 49.1 & 23.3 & 50.1 \\
& \nscl & 63.6 & 19.3 & 70.5 & 13.2 \\
& \msda & 2.4 & 90.5 & 2.4 & 85.8\\
\bottomrule
\end{tabular}
\captionof{table}{Error rates for positive and negative classes for \blse,
  \nscl, and \msda trained on the Amazon corpora and tested on the
  SemEval corpora.}
\label{table:erroranalysis}
\end{minipage}
\end{table}

Figure \ref{fig:lexicons} shows that the frequency-based lexicon gives
better results on the more divergent datasets, while the sentiment
lexicon performs slightly better on the similar datasets, but poorly
on the divergent datasets. The mutual information induced pivot
lexicons provide good results on all but the SemEval 2013
dataset. This is likely because the lexicon is too small (103 tokens)
to give a good mapping.

\subsection{Analysis of $M'$}
\label{analysism}
\begin{figure*}
\centering
\includegraphics[width=3.5in]{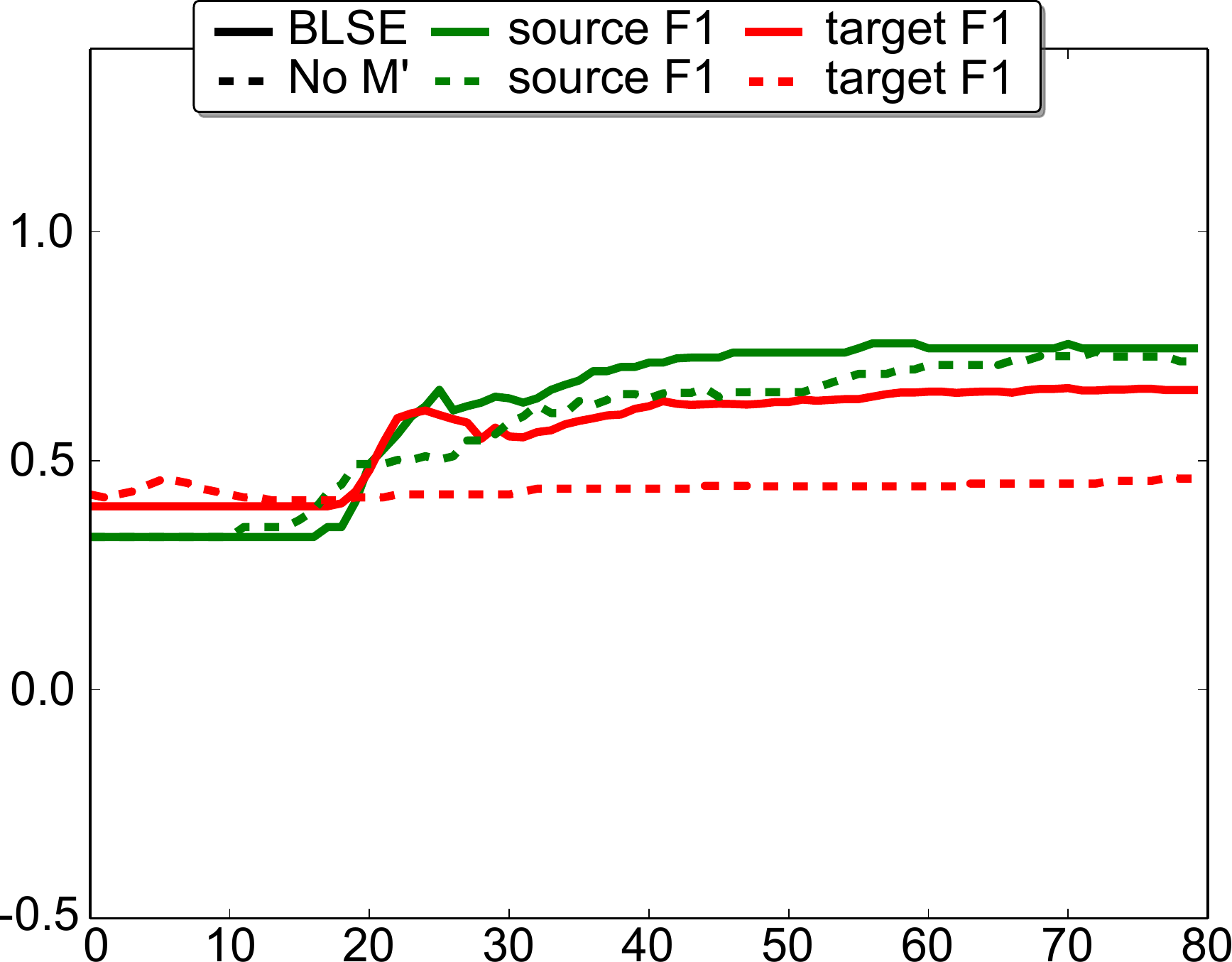}
\caption{The \blse model (solid lines) compared to a variant without the target domain
projection matrix $M'$ (dashed lines). The red green lines show \F on the source books dataset,
while the green lines show \F on the target SemEval 2013 dataset. The modified model does
not learn to predict sentiment on the target dataset (dashed red line).}
\label{fig:nomprime}
\end{figure*}

The main motivation for using two projection matrices $M$ and $M'$ is
to allow the original embeddings to remain stable, while the
projection matrices have the flexibility to align translations and
separate these into distinct sentiment subspaces. To justify this
design decision empirically, we perform an experiment to evaluate the
actual need for the target language projection matrix $M'$: We create
a simplified version of our model without $M'$, using $M$ to project
from the source to target and then $P$ to classify sentiment.

The results of this model are shown in
Figure~\ref{fig:nomprime}. The modified model does learn to predict
in the source language, but not in the target language. This confirms
that $M'$ is necessary to transfer sentiment in our model.

\section{Conclusion and Future Work}
We have presented an approach to domain adaptation which learns to
project mono-domain embeddings to a bi-domain space and use this
bi-domain representation to predict sentiment. We have experimented
with 20 domain pairs and shown that for highly divergent domains, our
model shows substantial improvement over state-of-the-art methods. Our
model constitutes a novel state of the art on 11 of the 20 domain
pairs.

One of the main advantages of this approach is that the learned
classifier can be used to classify sentiment in either of the two
domains without further tuning. In the future, we would like to extend
this model to learn multiple domain mappings at a time, effectively
permitting zero-shot domain adaptation at a large scale. This would
enable a single model to predict sentiment for a number of 
domains.

Another promising avenue for improvement is to create lexicons that
map concepts from the source domain to the target domain, \ie,
``read'' in the books domain to ``watch'' in the movies domain. It
would be interesting to see if it is possible to use vector algebra
\cite{Mikolov2013} to find similar concepts in different domains, \eg,
\textit{read - books + DVD = watch}. It would also be beneficial to
map multiword units across domains, \eg, ``not particularly exciting''
in DVD to ``not very reliable'' in electronics. This could be
particularly helpful for moving beyond a binary view of sentiment at
document-level, where domain adaptation would be of particular use,
given that the cost of annotation is higher for multi-class,
sentence-, or aspect-level classification.

A current disadvantage of our model might be that it uses skip-gram
embeddings trained on more than one domain. Therefore, it it would be
of interest to investigate if methods which create domain specific
embeddings \cite{He2011,Bollegala2014,Bollegala2015} are able to give
better results within our framework.
  
\section*{Acknowledgements}
We thank Manex Agirrezabal for proofreading the Basque abstract.

\bibliographystyle{acl}
\bibliography{acl2018}

\end{document}